%% file: preprint.tex
\documentclass[breaklinks=true]{article}

\usepackage[preprint]{neurips_2026}

\usepackage[utf8]{inputenc}
\usepackage[T1]{fontenc}
\usepackage{url}
\usepackage{caption}
\usepackage{hyperref}
\hypersetup{
    colorlinks,
    linkcolor={red!50!black},
    citecolor={blue!50!black},
    urlcolor={blue!80!black}
}
\usepackage{booktabs}
\usepackage{amsfonts}
\usepackage{microtype}
\usepackage{xcolor}

\usepackage{wrapfig}

\usepackage{amsmath,amssymb,bm}
\usepackage{mathtools}
\usepackage{amsthm}

\usepackage{graphicx}
\usepackage{pgfplots}
\usepackage{multirow}
\usepackage{longtable}
\usepackage{float}
\pgfplotsset{compat=1.18}
\usetikzlibrary{arrows.meta}
\usepackage{enumitem}

\usepackage{cleveref}

\theoremstyle{plain}

\theoremstyle{definition}

\theoremstyle{remark}

\title{A Study on Hidden Layer Distillation for Large Language Model Pre-Training}

\author{
  Maxime Guigon \\
  Google DeepMind\\
  \texttt{guimax@google.com} \\
  \And
  Lucas Dixon \\
  Google DeepMind \\
  \texttt{ldixon@google.com} \\
    \And
    Micha\"{e}l E. Sander \\
  Google DeepMind \\
\texttt{michaelsander@google.com} \\
}

\begin{document}

\maketitle
\begin{abstract}
Knowledge Distillation (KD) is a critical tool for training Large Language Models (LLMs), yet the majority of research focuses on approaches that rely solely on output logits, neglecting semantic information in the teacher's intermediate representations.
While Hidden Layer Distillation (HLD) showed potential for encoder architectures, its application to decoder-only pre-training at scale remains largely unexplored.
Through compute-controlled experiments, we benchmark HLD against logit-based KD and self-supervised baselines with Gemma3 3.4B as teacher and 123M and 735M students trained on up to 168B tokens from the C4 dataset. Our experiments show that HLD does not consistently outperform standard KD on downstream evaluation tasks.
Nevertheless, we show that HLD can yield a systematic perplexity gain over KD across all shared-hyperparameter configurations, suggesting that a latent signal can be extracted, but a breakthrough may be needed for it to play a more significant role in LLM pre-training.
\end{abstract}

\section{Introduction}

Large Language Models (LLMs) have achieved remarkable success across numerous Natural Language Processing (NLP) tasks, a phenomenon largely driven by the empirical observation of ``scaling laws'', which link model size and number of training tokens to performance gains \citep{chinchilla, sl1, sl2}.
However, this trend of continuously increasing scale introduces significant practical challenges, including soaring deployment costs, increased inference latency, and substantial energy consumption.
Reducing the computational footprint of LLMs, particularly during inference, has therefore become a critical area of research.

This paper focuses on Knowledge Distillation (KD) \citep{kd}, a robust and widely-adopted paradigm for transferring knowledge from a large, powerful ``teacher'' model to a smaller, more efficient ``student'' model by training the latter to reproduce the teacher's behavior.
KD has proven to be an enduring and highly effective technique for model compression across a decade of research, demonstrating utility in computer vision \citep{kdappvision1}, speech processing \citep{kdappspeech1}, and natural language processing for diverse architectures, including encoder-only LLMs \citep{distilBert} and various modern decoder-only LLMs spanning both pre-training and post-training stages \citep{gemma3,deepseekv3}.
Nevertheless, current distillation methods struggle to close the performance gap between students and teachers \citep{pkd,distilBert,minilm,mobilLLM}, yet teacher models frequently exhibit internal redundancy \citep{redudancy1,redudancy2} that foresees lossless compression.
Furthermore, academic exploration of distillation techniques within the context of massive, cost-intensive LLM pre-training remains limited, mainly exploring classic, logits-level KD \citep{pkd, kdscalinglaw}.

Hidden Layer Distillation (HLD), an encouraging early extension to KD introduced for CNNs in \citep{fitnets} (FitNets), which leverages a teacher's hidden states, has been largely applied to smaller-scale encoder(-decoder) language models showing potential advantage compared to KD 
\citep{tinybert,lessismore,distilBert,minilm,minilm2,layermatchingdontmatter,moebert,cka,patientkdbert,homobert}.
Despite this potential, the application of HLD has never been scaled to the magnitude of current state-of-the-art LLMs.

Motivated by this, we propose an exploration of HLD specifically for decoder-only models during the pre-training phase.
Evaluating LLMs is sensitive to minor hyperparameter changes that can disproportionately affect final performances; we therefore establish rigorous baselines ensuring that any performance gains are attributable to the evaluated distillation methodology rather than artifacts.
Our experiments are built upon the open-source NanoDo codebase \citep{nanodo}, the C4 dataset \citep{C4}, and open-weight models from the Gemma family \citep{gemma3}.

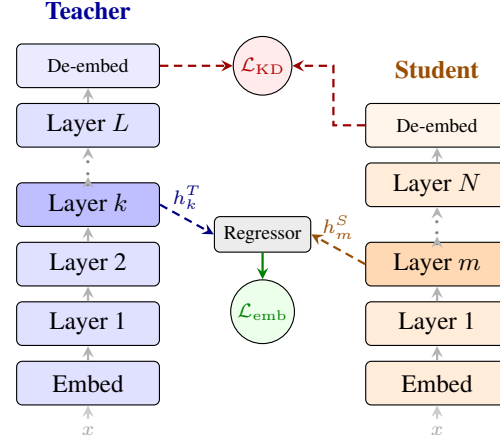
\begin{wrapfigure}[21]{H}{0.48\textwidth}
\vspace{-10pt} 
\resizebox{\linewidth}{!}{
\begin{tikzpicture}[
    block/.style={draw, rounded corners=2pt, minimum width=1.8cm, minimum height=0.55cm,
                  font=\small},
    teacher/.style={block, fill=blue!12},
    student/.style={block, fill=orange!15},
    reg/.style={draw, rounded corners=2pt, fill=gray!15, minimum width=1.2cm,
                minimum height=0.4cm, font=\scriptsize},
    loss/.style={draw, circle, inner sep=1.5pt, font=\scriptsize\bfseries, fill=white},
    arrow/.style={->, >=stealth, thick},
    dasharrow/.style={->, >=stealth, thick, densely dashed},
]

\node[font=\small\bfseries, text=blue!60!black] at (-2.2, 4.7) {Teacher};
\node[teacher] (t_emb) at (-2.2, 0)    {Embed};
\node[teacher] (t1)    at (-2.2, 0.75) {Layer 1};
\node[teacher] (t2)    at (-2.2, 1.5)  {Layer 2};
\node[teacher, fill=blue!25] (tk) at (-2.2, 2.25) {Layer $k$};
\node[font=\normalsize, text=gray] at (-2.2, 2.75) {$\vdots$};
\node[teacher] (tL)    at (-2.2, 3.25) {Layer $L$};

\node[font=\small\bfseries, text=orange!60!black] at (2.2, 3.95) {Student};
\node[student] (s_emb) at (2.2, 0)    {Embed};
\node[student] (s1)    at (2.2, 0.75) {Layer 1};
\node[student, fill=orange!30] (sm) at (2.2, 1.5) {Layer $m$};
\node[font=\normalsize, text=gray] at (2.2, 2.0) {$\vdots$};
\node[student] (sN)    at (2.2, 2.5) {Layer $N$};

\draw[arrow, gray!60] (t_emb) -- (t1);
\draw[arrow, gray!60] (t1) -- (t2);
\draw[arrow, gray!60] (t2) -- (tk);
\draw[arrow, gray!60] (tk.north) -- (-2.2, 2.6);
\draw[arrow, gray!60] (-2.2, 2.9) -- (tL.south);

\draw[arrow, gray!60] (s_emb) -- (s1);
\draw[arrow, gray!60] (s1) -- (sm);
\draw[arrow, gray!60] (sm.north) -- (2.2, 1.85);
\draw[arrow, gray!60] (2.2, 2.15) -- (sN.south);

\node[font=\scriptsize, text=gray!70] at (-2.2, -0.6) {$x$};
\node[font=\scriptsize, text=gray!70] at (2.2, -0.6) {$x$};
\draw[arrow, gray!40] (-2.2, -0.45) -- (t_emb);
\draw[arrow, gray!40] (2.2, -0.45) -- (s_emb);

\node[block, fill=blue!8]   (t_head) at (-2.2, 4.0) {\scriptsize De-embed};
\node[block, fill=orange!8] (s_head) at (2.2, 3.25) {\scriptsize De-embed};
\draw[arrow, gray!60] (tL) -- (t_head);
\draw[arrow, gray!60] (sN) -- (s_head);

\node[loss, fill=red!8] (lkd) at (0, 4.0) {\color{red!70!black}$\mathcal{L}_{\mathrm{KD}}$};
\draw[dasharrow, red!60!black] (t_head.east) -- (lkd);
\draw[dasharrow, red!60!black] (s_head.west) -- +(-0.4,0) |- (lkd);

\node[reg] (R) at (0, 1.875) {Regressor};
\node[loss, fill=green!8] (lhld) at (0, 0.9) {\color{green!50!black}$\mathcal{L}_{\mathrm{emb}}$};

\draw[dasharrow, blue!50!black] (tk.east) -- (R.west)
    node[midway, above, font=\scriptsize, text=blue!50!black] {$h_k^T$};
\draw[dasharrow, orange!60!black] (sm.west) -- (R.east)
    node[midway, above, font=\scriptsize, text=orange!60!black] {$h_m^S$};
\draw[arrow, green!50!black] (R) -- (lhld);

\end{tikzpicture}
}
\caption{Overview of HLD. The student receives two training signals: $\mathcal{L}_{\mathrm{KD}}$ matches the teacher's output logits, while $\mathcal{L}_{\mathrm{emb}}$ aligns a student hidden state $h_m^S$ with a teacher hidden state $h_k^T$ through a learned regressor.}
\label{fig:hld_overview}
\end{wrapfigure}

More precisely, we make the following contributions:
\begin{itemize}[leftmargin=1.5em,itemsep=2pt,topsep=2pt]
\item We provide a systematic evaluation of HLD applied specifically to the pre-training of decoder-only LLMs by training 123M and 735M gemma students models \citep{gemma3} on 8B and 168B tokens with a gemma 3.4B teacher.    
\item We establish a rigorous evaluation framework that benchmarks methods based on strict computational equivalence (FLOPs) rather than token count, explicitly accounting for the non-negligible cost of de-embedding and loss computation in sub-billion parameter students.
\item We demonstrate through extensive compute-matched experiments and evaluation over Wikitext~103, HellaSwag, WinoGrande, LAMBADA, PIQA, and ARC-E that, compared to KD:
\begin{itemize}[leftmargin=1.5em,itemsep=0pt,topsep=0pt]
    \item (i) joint optimization of hidden layer alignment and logit distillation (HLDC) performs on par with KD;
\end{itemize} 
\end{itemize}

\begin{itemize}[leftmargin=30pt,itemsep=0pt,topsep=0pt,label=\textbf{--}]
    \item (ii) sequential optimization (HLDF) yields modest improvements in C4 perplexity across all shared-hyperparameter configurations, while performance across downstream evaluation remains comparable.
\end{itemize}

\section{Background and Related Works}

\paragraph{Transformers.}
The Transformer \citep{transformer} is the main neural network architecture for the modeling of sequential data.
Current generative LLMs use the decoder-only version.
Given an input string $s$, the tokenizer produces a sequence of discrete token indices $\mathbf{T} = [t_1, \dots, t_n] \in \{0, \dots, V-1\}^n$, where $V$ is the vocabulary size.
These are mapped to continuous embeddings $\mathbf{H}^0 = [\mathbf{h}_1^0, \dots, \mathbf{h}_n^0] \in \mathbb{R}^{d_{emb} \times n}$.
The backbone consists of $D$ residual layers, with update rule $\mathbf{H}^{k+1} = \mathbf{H}^k + f_{layer}^{k+1}(\mathbf{H}^k)$.
Finally, a de-embedding layer projects the final representations $\mathbf{H}^D$ to logits $\mathbf{Z} \in \mathbb{R}^{V \times n}$.
The probability distribution for the next token is $\mathbf{p}_n = \text{softmax}(\mathbf{z}_n / \tau)$, where $\tau$ is the temperature parameter.

\paragraph{Self-Supervised Training.}
In the absence of a teacher model, the conventional approach for training a Transformer involves self-supervised learning on a comprehensive text corpus. 
For decoder-only models designed for generative tasks, the optimization objective typically employs a causal language modeling loss.
If $\bm{\theta}$ represents the model parameters, this loss is defined for a single input sequence as:
\begin{equation}
\mathcal{L}_{\text{data}}(\bm{\theta}) = -\frac{1}{n-1}\sum_{i=1}^{n-1} \log \mathbf{p}_i[t_{i+1}].
\label{eq:data}
\end{equation}
We refer to this objective as NLL (Negative Log-Likelihood).

\paragraph{Knowledge Distillation.}
In contexts where a teacher model is available, the Knowledge Distillation (KD) \citep{kd} framework can be applied to any classification model. 
This optimization objective is designed to align the student model's output distribution with that of the teacher. 
To quantify the discrepancy between these distributions (denoted here as $p$ and $q$) and calculate the associated loss, the Kullback-Leibler (KL) divergence is commonly utilized with a temperature term within the softmax function: 
\[
\operatorname{KL}(p \,\|\, q) = \sum_{x \in \mathcal{X}} p(x) \log \frac{p(x)}{q(x)}.
\]
This approach is recognized for its superior performance compared to NLL, as the teacher's logits encapsulate more comprehensive information than one-hot ground-truth labels. Denoting the output probabilities of the student as $\mathbf{p}_i^S$ and the teacher as $\mathbf{p}_i^T$, the associated objective is defined as follows:
\[
\mathcal{L}_{\text{logits}}(\bm{\theta}_S) = \frac{\tau^2}{n}\sum_{i=1}^{n}\text{KL}(\mathbf{p}_i^T || \mathbf{p}_i^S).
\]
In practice, the loss used is a composite of the NLL loss and the knowledge distillation loss:
\begin{equation}
\mathcal{L}_{\text{KD}}(\bm{\theta}_S) = (1-\alpha)\mathcal{L}_{\text{data}}(\bm{\theta}_S) + \alpha\mathcal{L}_{\text{logits}}(\bm{\theta}_S),
\label{eq:kd}
\end{equation}

\paragraph{Distillation in open-source research.}
The open-source literature on distillation techniques for decoder-only models is relatively sparse regarding large scale models. \citet{pkd} make a similar observation, motivating their thorough exploration of the design space using transparent, open-weight teachers of 9B and 32B parameters and student scales ranging from 330M to 6.8B parameters. 
They observe that knowledge distillation consistently yields improvements compared to NLL and explore the hyperparameter space to identify optimal configurations. 
Similarly, \citet{kdscalinglaw} propose a distillation scaling law to predict the performance of a distilled model based on its compute budget and the allocation between student and teacher models. 
They scale teachers and students from 143M to 12.6B parameters, trained on up to 16 times the Chinchilla-optimal token budget \citep{chinchilla}.

\paragraph{Distillation in proprietary development.}
Parallel to these academic efforts, major industry laboratories have widely adopted distillation for the training of state-of-the-art models \citep{qwen3,gemini25,gemma3,deepseekv3}. 
While the specific training pipelines and hyperparameters often remain proprietary, the high performance of these releases—some of which are open-weights models—serves as strong empirical evidence for the efficacy of distillation in the pre-training phase.

\paragraph{Hidden Layer Distillation.} 
Introduced by \citet{fitnets} (FitNets), HLD employs a two-phase training strategy. The first phase aligns the student's intermediate hidden activations with the teacher's, using a learned regressor. The second phase then applies standard knowledge distillation across the full architecture.

\paragraph{HLD for Transformers.}
Contemporary research into extracting value from the latent representations of Transformer models has introduced various modifications to the initial framework, resulting in a diverse array of loss functions. Examples of these experimented variations include the alignment of attention maps \citep{tinybert} and value-relation \citep{minilm, minilm2}, multi-layer matching strategies using distinct layer allocation schemes \citep{layermatchingdontmatter}, the prior training of task-aware filters to match the student and teacher hidden dim \citep{lessismore}, the application of non-learned transformation-invariant techniques to reconcile disparate hidden activation dimensions \citep{cka},  direct RMS matching \citep{patientkdbert,moebert} or cosine distance \citep{distilBert} of normalized activations for teacher and student models of identical width.

Notably, current implementations typically favor simple linear regressors over the multi-layer models originally proposed by \citet{fitnets}. Furthermore, the two-phase training protocol described in earlier literature is frequently replaced by a single-phase process that minimizes an aggregated objective function:
\begin{equation}
\mathcal{L}_{HLD} = \beta\mathcal{L}_{data} + \alpha\mathcal{L}_{logits} + \gamma \mathcal{L}_{emb}.
\label{eq:hld}
\end{equation}
Unfortunately, most existing studies focus on the post-training phase and on encoder or encoder-decoder models. To our knowledge, \citet{lessismore} are the only ones to experiment with HLD on decoder-only models and only with continual pre-training. They worked with \citet{gpt2}'s GPT-2$_{12}$ (120M parameters) as the teacher and GPT-2$_{6}$ (82M parameters) as the student, which is initialized with a subset of the teacher's layers.
To the best of our knowledge, our work is the first open-source study of HLD during pre-training for causal LLMs.

\section{Proposed Method}
To investigate both the original HLD formulations and contemporary approaches in the Transformer literature, we evaluate the following two methods.  
\paragraph{Sequential Optimization (HLDF).}
We adapt \citet{fitnets}'s two-phase protocol as a minimal extension to standard distillation that isolates the value of intermediate feature guidance.
The regressor $f_{reg}(.;\bm{\theta_R})$ is a one-layer dense perceptron.
Phase 1 minimizes a normalized hint-training (HT) loss:
\begin{equation}
\mathcal{L}_{HT}(\bm{\theta_S},\bm{\theta_R})  = \text{MeanSquaredError}\!\left(\frac{\mathbf{H^{D_T/2}_T}}{||\mathbf{H^{D_T/2}_T}||},\frac{f_{reg}(\mathbf{H^{D_S/2}_S};\bm{\theta_R})}{||f_{reg}(\mathbf{H^{D_S/2}_S};\bm{\theta_R})||}\right)\!.
\label{eq:ht}
\end{equation}
Phase 2 runs standard KD with $\mathcal{L}_{KD}$ as objective.
In short:
\begin{itemize}
    \item 
        \textbf{Phase 1}: The objective is $\mathcal{L}_{HT}(\bm{\theta_S},\bm{\theta_R})$
    \item 
        \textbf{Phase 2}: The objective is $\mathcal{L}_{KD}(\bm{\theta_S})$
        
\end{itemize}
We use HLDF (F stands for FitNet) to refer to this training method.

\paragraph{Joint Optimization (HLDC).}
To align with loss functions previously explored in the literature for decoder architectures, we also implement a single-stage training using $\mathcal{L}_{\text{HLD}}$ \eqref{eq:hld}, with the embedding loss defined as:
\[
\mathcal{L}_{emb}(\bm{\theta_S},\bm{\theta_R})  = \text{MeanSquaredError}\!\left(\frac{\mathbf{H^{D_T/2}_T}}{||\mathbf{H^{D_T/2}_T}||}, \frac{\mathbf{W_R}\mathbf{H^{D_S/2}_S}}{||\mathbf{W_R}\mathbf{H^{D_S/2}_S}||}\right)\!.
\]
We use HLDC (C stands for composite) to refer to this training method.

\paragraph{Normalization Strategy.}
Unlike FitNets, we compare activations after normalization, consistent with prior work \citep{distilBert, patientkdbert}.
This approach is essential for Gemma 3 architectures in which we observed that the unnormalized residual stream can grow significantly in magnitude across layers.
Since the student model possesses fewer layers than the teacher, it cannot naturally match the teacher's internal activation scale.
For instance, the mean squared magnitude of the teacher ($\mathbf{H^{D_T/2}_T}$) reaches approximately $3 \times 10^5$, whereas the pre-trained Gemma3 270M student ($\mathbf{H^{D_S/2}_S}$) only reaches $\approx 1 \times 10^5$.

\paragraph{Data.}
We train the models on the English subset of the C4 dataset \citep{C4}.
This is a well-established corpus, also used by \citet{kdscalinglaw}. With around 150B tokens, it is sufficiently large to ensure that few sample are seen more than once during training.

\paragraph{Teacher model.}
We pretrain a Gemma3 4B \citep{gemma3} as teacher on 106B unique tokens with an alternative SentencePiece unigram model with a vocabulary size of 32k (making it a 3.4B model). This choice is motivated by the large original Gemma vocabulary (256k tokens), which is multilingual while we train exclusively on English data.
Gemma3 4B is a 34-layer pretrained decoder-only Transformer with an embedding dimension $d_T=2560$.

\paragraph{Student models.}
As student models, we use Gemma3 270M and 1B \citep{gemma3} with standard random initialization.
They are respectively 18 and 26-layer decoder-only Transformers sharing the architecture pattern of the teacher, with embedding dimensions $d_S$ of 640 and 1152.
With the 32k vocabulary, they become 123M ($\sim$100M backbone) and 735M ($\sim$700M backbone) total parameters, yielding 27:1 and 4.5:1 compression ratios.
The investigation of alternative teacher-student pairs and different compression ratios remains a priority for future research.

\paragraph{Logits.}
A common industry practice for optimizing offline storage is to retain only the top-$k$ logits, where $k$ is significantly smaller than the total vocabulary size; this method reduces storage requirements by several orders of magnitude. For instance, the Gemma 3 tokenizer possesses a vocabulary of approximately 256,000 tokens, and a standard value of $k = 128$ is typically employed. This threshold also functions as a regularizer, shielding the student model from the noise inherent in low-probability logits. In this study, $k$ is fixed at 128 and is not subject to further analysis, as \citet{pkd} observe that the choice of $k$ makes little difference.

\paragraph{Intermediate activations.}
However, the application of HLD necessitates the storage of full activations as vectors of the teacher embedding space, preventing the use of the top-$k$ truncation technique.
In order to fit storage constraints, minimize the number of hyperparameters, we consider that evaluating the efficacy of HLD using the activations from a single teacher layer presents a significant challenge and a robust initial step for investigating hidden layer distillation.
To stay align with \citet{fitnets}, we select the median layer of the teacher model  for data retention and subsequent distillation.

\paragraph{Optimization.}
Our experiments are built upon the NanoDo codebase \citep{nanodo}, an open-source vanilla decoder-only pre-training code base. 
We use the AdamW optimizer \citep{adamw} with $\beta_1 = 0.9$, $\beta_2 = 0.98$, $\epsilon = 10^{-9}$, a weight decay of $0.1$, a batch size of 256 and a context length of 2048.
The learning rate follows a WSD schedule \citep{wsd} with 1{,}000 warmup steps and a 10\% decay phase ending at $\eta/100$.

\paragraph{Hardware.}
All experiments, including the storage of the teacher's logits and activations, as well as the training and evaluation of the models, are conducted on TPUv4 and TPUv7 chips.

\paragraph{Evaluation.}
Since our study focuses exclusively on the pre-training phase, we restricted performance assessments to perplexity or score-based evaluations across: \textbf{Wikitext~103} \citep{wikitext}, \textbf{HellaSwag} \citep{hellaswag}, \textbf{WinoGrande} \citep{winogrande}, \textbf{LAMBADA} \citep{lambada}, \textbf{PIQA} \citep{piqa}, \textbf{ARC-E} \citep{arc}, and a held-out validation split of \textbf{C4} \citep{C4}.

\section{Experiments}

\paragraph{Hyperparameter Considerations.}
Evaluating the efficacy of a novel distillation methodology is complex, as subtle hyperparameter sensitivities can disproportionately influence both convergence and terminal performance.
A fair comparison must verify that the baseline is well-tuned: would a practitioner be better served by adopting the new method or simply by performing a learning rate sweep on conventional KD?
Rather than tuning HLD to beat KD, we characterize its general behavior over the peak learning rates $\eta_{KD}$ and $\eta_{HT}$, the temperature $\tau$, and the NLL weight $\alpha$ in the global objective \eqref{eq:kd} and \eqref{eq:hld}.

\paragraph{Compute-matched comparisons.}
Following the establishment of scaling laws for LLMs \citep{chinchilla}, compute-matched comparisons have become a methodological necessity.
Consequently, \textbf{all experimental evaluations in this work are conducted between models trained using equivalent computational budgets}.

\paragraph{Overtraining Unit.}
We quantify computational expenditure using an ``overtraining unit'', denoted as $\text{OT}_k$. This unit represents $k$ times the compute required to reach the Chinchilla-optimal point for our student model with NLL \citep{chinchilla}. 
For instance, $\text{OT}_1$ corresponds to the budget required to optimally train the 123M-parameter student on 2B tokens ($20 \times \text{backbone parameters}$). However, \citet{chinchilla}'s compute-optimal models do not account for inference costs during the model's serving lifetime. 
A newer paradigm, \textit{overtraining}, suggests training models well beyond compute optimality to amortize serving costs \citep{overtraining}. 
To maintain relevance within this paradigm we train:
\begin{itemize}
    \item the 123M student up to $\text{OT}_{54}$ (108B tokens) 
    \item the 735M student up to $\text{OT}_{12}$ (168B tokens). 
\end{itemize}

\paragraph{FLOPs accounting.}
We refine the standard $6ND$ approximation for training compute, as the conventional formula neglects embedding and de-embedding layers (\cref{tab:cost}).
While these layers are negligible in large-scale Transformers, they represent a significant portion of the total FLOPs for our student models due to the large vocabulary size ($V = 32{,}000$).
The backward pass cost is equal to the forward pass for non-parametric layers and twice the forward pass for layers with learnable parameters (see \textit{e.g.} \citep{thebible}).
Let $N$ denote the number of student backbone parameters, $d_S$ the internal dimension, $d_T$ the teacher's hidden dimension, and $N_{\text{reg}}$ the parameters in the mapping regressor.
Following \citet{scalingbook} and \cref{tab:cost}, the estimated compute cost $\mathcal{C}_{data}$ per token of self-supervised learning \eqref{eq:data}, the cost $\mathcal{C}_{KD}$ of conventional KD \eqref{eq:kd}, the cost $\mathcal{C}_{HT}$ of the first phase of the HLDF that train only up to the matched layer \eqref{eq:ht} and the cost $\mathcal{C}_{HLDC}$ of HLDC \eqref{eq:hld} are:
\begin{align*}
&\mathcal{C}_{data} \approx \underbrace{6N}_{\text{backbone}} + \underbrace{6d_{S}V}_{\text{logits}}, \quad \mathcal{C}_{KD}   \approx \mathcal{C}_{data} + \mathcal{C}_{\text{Teacher}}, \\
&\mathcal{C}_{HT}   \approx \underbrace{3N}_{\text{half backbone}} + \underbrace{6 N_{reg}}_{\text{Mapping}} +\mathcal{C}_{\text{Teacher}}/2 \quad \text{and} \quad \mathcal{C}_{HLDC} \approx \mathcal{C}_{KD}+ \underbrace{6 N_{reg}}_{\text{Mapping}}
\end{align*}
Since teacher logits are already pre-computed, $\mathcal{C}_{\text{Teacher}}$ is negligible.
For the 123M student, the relative costs are 
\[
\frac{\mathcal{C}_{KD}}{\mathcal{C}_{data}} \approx 1.000, \quad \frac{\mathcal{C}_{HT}}{\mathcal{C}_{data}} \approx 0.442, \quad \frac{\mathcal{C}_{HLDC}}{\mathcal{C}_{data}}\approx 1.027.
\]
Therefore, while Phase~1 of HLDF updates approximately half the model's backbone parameters, its computational cost is significantly less than 50\% of the KD baseline.
This efficiency is amplified by the high ratio of vocabulary size to model parameters in the 123M architecture, where the bypassed output projection accounts for a large fraction of FLOPs.
As model size scales and backbone computations dominate, this advantage vanishes: a Gemma3 27B student \citep{gemma3} with a $2\times$ wider teacher yields $\frac{\mathcal{C}_{HT}}{\mathcal{C}_{data}} \approx 0.5001$.

\paragraph{HLD-specific hyperparameters.}
To emulate a ``no-tuning'' scenario, we fix HLDF-specific hyperparameters to heuristic values.
We employ a single-layer MLP regressor with an expansion factor of 4 (relative to the student's hidden dimension), mirroring the standard architecture of the student's internal feed-forward layers.
For the Phase~1 schedule, we use a constant learning rate $\eta_{HT}$ following a 1{,}000-step linear warmup, with no subsequent decay.
For HLDC, the only exclusive hyperparameter is $\gamma$, the weight of the embedding loss.

\paragraph{Two evaluation methods.}
We compare HLD against KD in two complementary ways.
First (\S\ref{average}), we perform pointwise comparisons at shared hyperparameter configurations, evaluating robustness across a broad grid without any method-specific tuning.
Second (\S\ref{ft}), we give each method its independently selected best hyperparameters, enabling a ceiling-to-ceiling comparison.
\input{tab/pahp_transposed.tex}

\paragraph{Shared hyperparameters.}
For each set of hyperparameters $(\eta = \eta_{KD} = \eta_{HT}, \tau, \alpha)$, we run one NLL training and one KD training as baselines.
For HLDF, we use $\eta_{HT} = \eta$ in Phase~1 and $(\eta, \tau, \alpha)$ in Phase~2, varying the budget split $P_1$ between phases.
For HLDC, we explore small values of $\gamma$ \citep{lessismore} with the same $(\eta, \tau, \alpha)$ as the baseline.
We select values for $\eta, \tau,$ and $\alpha$ from regimes where KD is known to perform well \citep{pkd}, adapting $\eta$ to our batch size via square-root scaling \citep{batchsize1}.
The complete set of explored hyperparameters is summarized in \cref{hp}.
To evaluate the general performance of HLD over KD, 
we perform a pointwise comparison using identical hyperparameter sets $(\eta, \tau, \alpha)$.
Detailed results are provided in \cref{fig:c4_ppl,fig:downstream_735,fig:downstream_123,fig:scatter_ppl,fig:scatter_123,fig:scatter_735}.
\label{average}

\paragraph{Best hyperparameters.}
We ask whether HLD outperforms KD when each method uses its best configuration. We select for each method the best-performing $(\eta^\ast,\tau^\ast,\alpha^\ast)$ from the shared-hyperparameter grid \cref{hp}, and compare the resulting peak performances head-to-head in \cref{tab:protocolB_combined}.
\label{ft}

\section{Results}

We present our results in four complementary views:
\begin{itemize}
    \item \Cref{fig:c4_ppl,fig:downstream_123,fig:downstream_735} show histograms of the improvement of HLD over KD at fixed hyperparameter sets, illustrating the distribution of gains of each evaluated method against the KD baseline.
    \item \Cref{fig:scatter_ppl,fig:scatter_123,fig:scatter_735} in the appendix are scatter plots in which each point corresponds to one shared hyperparameter set: the $x$-axis reports the KD score and the $y$-axis the score of the compared method (HLD or NLL). These plots reveal the general behavior of each method relative to its baseline, as well as the behavior of KD relative to its NLL baseline. 
    \item \Cref{tab:protocolB_combined} reports the best score achieved by each method, alongside the teacher and the NLL baseline. 
    \item Finally, \Cref{tab:xps_123,tab:xps_735} in the appendix list the full set of training runs together with their evaluation results.
\end{itemize}

\input{plot/ppl_horizontal.tex}
\input{tab/protocoleB_combined.tex}

\paragraph{Baseline.}
A prerequisite for any meaningful comparison is that the KD baseline itself is sufficiently well-tuned to serve as a competitive reference for HLD. Several elements support this. First, as shown in \cref{tab:protocolB_combined}, our best KD configuration outperforms NLL by a clear margin on every benchmark, even on noisier evaluations such as WinoGrande. Second, on the perplexity scatter plots (\cref{fig:scatter_ppl}), all but one KD run beats NLL for the 735M student, and this trend is corroborated by the downstream evaluations (\cref{fig:scatter_735}). The picture is more mixed for the 123M student: roughly half of the KD runs fail to beat NLL on C4 perplexity (\cref{fig:scatter_ppl}), but the downstream results (\cref{fig:scatter_123}) again show KD outperforming NLL on nearly all evaluation points. Taken together, these observations indicate that our KD baseline is genuinely competitive, and that any comparison with HLD is being made against a strong reference rather than a weak one—a prerequisite that, in our knowledge, is not always met in the HLD literature.

\paragraph{Shared hyperparameters.}
On perplexity, HLDC performs on par with KD on C4, while HLDF achieves a modest but systematic improvement (\cref{fig:c4_ppl}). The effect size is small and, given the computational cost of pre-training, we lack the statistical power to claim significance from independent seeds; we instead rely on the consistency of the sign of the improvement across the entire shared-hyperparameter grid. Downstream evaluations tell a different story: as shown in \cref{fig:downstream_735,fig:downstream_123}, the score distributions are more spread out and roughly centered around zero on all benchmarks, indicating that the potential hidden-layer signal does not translate into consistent gains. A closer look at the perplexity scatter plot (\cref{fig:scatter_ppl}) further reveals that HLDF's improvement over KD occurs predominantly in regimes where KD itself performs poorly, while in regimes where KD is already efficient the gain becomes negligible.

\input{plot/downstream_735m.tex}
\input{plot/downstream_123m.tex}
\paragraph{Best hyperparameters.}
\Cref{tab:protocolB_combined} report the best-configuration performance for each method on both students. No method dominates KD across the seven evaluations: differences are small and split between methods, reinforcing the conclusion that, in our setup, HLD does not provide a consistent, reliable improvement over logit-based distillation that is tuned enough and under compute-matched conditions.

\paragraph{Reconciling with prior work.}
Our findings appear to contradict a body of work reporting consistent gains from intermediate-layer matching \citep{tinybert,distilBert,minilm,minilm2,patientkdbert,lessismore,moebert,cka}. We see three methodological factors that plausibly account for the discrepancy. 
(i) \emph{Compute accounting.} Most prior comparisons match token counts rather than FLOPs and some variations of HLD cost more by token than KD. 
(ii) \emph{Architecture and training phase.} The HLD literature is dominated by encoder and encoder-decoder models in post-training or task-specific distillation regimes, where representations are already shaped by a pre-training objective and the alignment target is correspondingly well-defined. Causal decoders trained from scratch present a fundamentally different optimization landscape, and our results suggest the inductive bias HLD provides is less valuable in this setting. 
(iii) \emph{Baseline tuning.} When HLD is introduced as the contribution, the KD baseline is rarely swept as carefully as the proposed method. Our shared-hyperparameter protocol explicitly avoids this asymmetry, and the resulting KD baseline is strong enough to absorb most of HLD's apparent advantage. 
(iv) \emph{Methodology.} Adapting the HLD methodology from few millions parameters CNN models to almost billions parameters transformers is not straightforward and would certainly need more profound design choices (losses, adapter architecture, protocol) to extract the information from the latent representations of the teacher.

\paragraph{On the residual HLDF signal.}
The systematic perplexity gain from HLDF is small but does not vanish under our protocol. One interpretation is that hint-training acts as warm-start mechanism — aligning the student's mid-network representations early may provide a better initialization for the subsequent KD phase, without fundamentally changing what the student can learn. This would explain why the effect appears in C4 perplexity (sensitive to fine-grained distributional fit) but not in downstream benchmarks (sensitive to coarser capabilities that converge similarly under both protocols) and why the difference with KD vanishes in regimes where KD performs the best.

\section*{Conclusion}

This work presents a compute-controlled evaluation of Hidden Layer Distillation (HLD) for the pre-training of causal LLMs on the English version of C4.
Despite the appeal of aligning intermediate representations to capture the teacher's deep internal knowledge, our results indicate that HLD does not consistently outperform a well-tuned logit-based KD baseline when training budgets are strictly equalized.
Furthermore, the complexity created by HLD introduces hyperparameter sensitivity that may outweigh its potential benefits in practical deployment scenarios.
These findings underscore the resilience of standard logit-based distillation and highlight the necessity of exact FLOPs accounting when evaluating novel supervision techniques.

That said, HLDF yields a systematic perplexity gain over KD, suggesting a latent signal can be extracted that (i) need to be confirmed through stronger statistical evidence and a broader context (larger scales, different compression ratios, alternative model families) and (ii) would likely require a significant breakthrough or substantial design choices (losses, adapter architecture, protocol) to thoroughly extract the information from the teacher's latent representations.

\clearpage

\bibliographystyle{abbrvnat}
\bibliography{biblio}

\clearpage

\appendix
\input{appendix.tex}

\end{document}

%% file: tab/pahp_transposed.tex
\begin{table*}[h!]
  \centering
  \caption{Hyperparameter grid.}
      \begin{tabular}{lccccc}
        \toprule
        Student size & $\eta$ & $\tau$ & $\alpha$ & $\gamma$ & $P_1$ \\
        \midrule
        123M & $\{1, 4, 16\} \times 10^{-4}$ & $\{0.5, 1\}$ & $\{0.7, 0.9\}$ & $\{0.1, 0.05\}$ & $\{1\%, 5\%\}$ \\
        735M & $\{1, 4, 16\} \times 10^{-4}$ & $\{0.5, 1\}$ & $\{0.7, 0.9\}$ & $\{0.1, 0.05\}$ & $\{1\%, 4\%\}$ \\
        \bottomrule
      \end{tabular}
  \label{hp}
\end{table*}

%% file: plot/ppl_horizontal.tex
\begin{figure*}[h!]
\centering
\includegraphics[width=\textwidth]{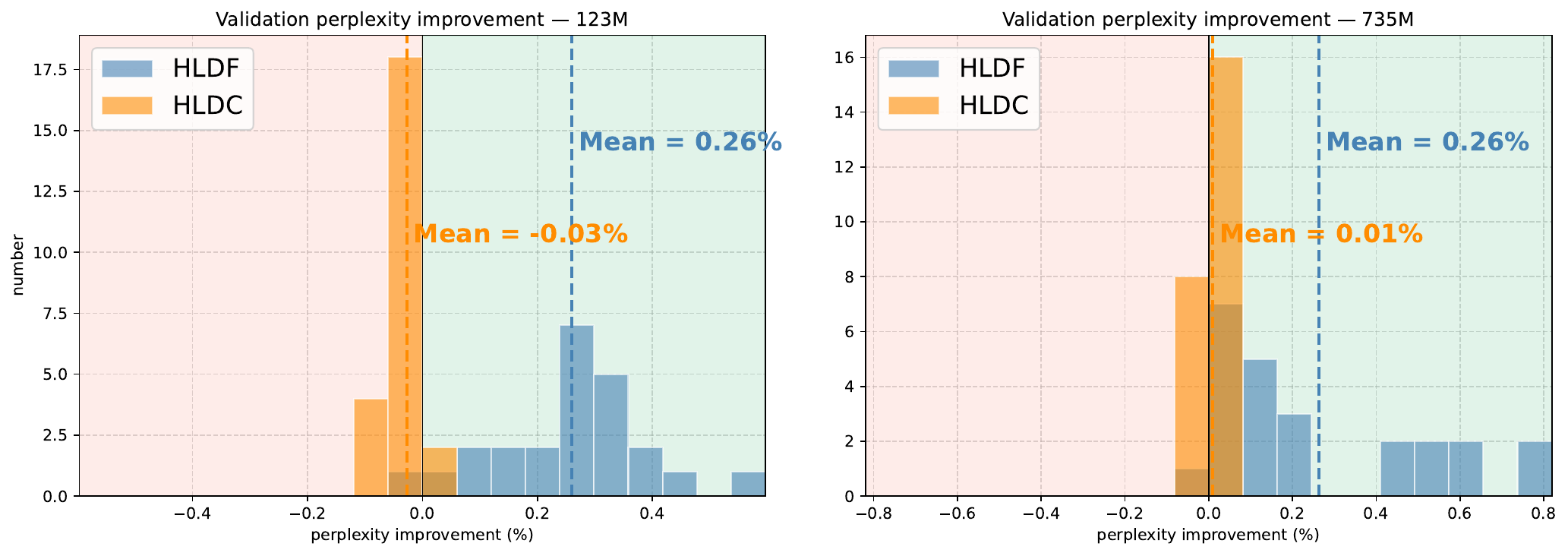}
\caption{
    C4 perplexity improvement over KD.
    Distribution of pointwise improvements across all hyperparameter configurations for both sizes.
    Positive values indicate lower perplexity than KD.
}
\label{fig:c4_ppl}
\end{figure*}

%% file: tab/protocoleB_combined.tex
\begin{table*}[h!]
    \centering 
    \small 
    \setlength{\tabcolsep}{4.5pt} 
    \caption{\textbf{Best Performance Comparison.} Test log-perplexity ($\downarrow$) or the error rate ($\downarrow$) for the optimal hyperparameters found within the compute-matched search space. Results are presented as \textbf{123M / 735M} student performances. \textbf{Bold} indicates the best result for a given student size.}
    \begin{tabular}{lccccccc}
        \toprule
        \textbf{Method} & \textbf{C4} & \textbf{Wikitext} & \textbf{HSwag} & \textbf{PIQA} & \textbf{ARC-E} & \textbf{WinoGd} & \textbf{LMBDA}\\
        \midrule
        Random  & --            & --            & 0.750         & 0.500         & 0.750         & 0.500         & $\approx$ 1.0 \\
        Teacher & 2.397         & 2.203         & 0.319         & 0.225         & 0.339         & 0.365         & 0.377         \\
        \midrule
        \multicolumn{8}{c}{\textit{Student Performances: 123M / 735M}} \\
        \midrule
        NLL     & 3.063 / 2.639 & 3.062 / 2.453 & .644 / .440 & .338 / .261 & .529 / .416 & .440 / .423 & .527 / .445 \\
        KD      & 3.005 / 2.609 & 2.984 / \textbf{2.406} & \textbf{.622} / .428 & \textbf{.321} / .256 & \textbf{.505} / \textbf{.402} & .433 / .407 & .513 / \textbf{.423} \\
        HLDC    & 3.005 / 2.608 & 2.938 / 2.438 & .630 / .430 & .323 / \textbf{.251} & \textbf{.505} / .415 & \textbf{.423} / .405 & \textbf{.505} / .425 \\
        HLDF    & \textbf{3.000} / \textbf{2.607} & \textbf{2.875} / 2.422 & .626 / \textbf{.426} & .322 / .261 & .519 / .409 & .437 / \textbf{.404} & .507/.428 \\
        \bottomrule
    \end{tabular}
    \label{tab:protocolB_combined}
\end{table*}

%% file: plot/downstream_735m.tex
\begin{figure*}[h!]
\centering
\includegraphics[width=\textwidth]{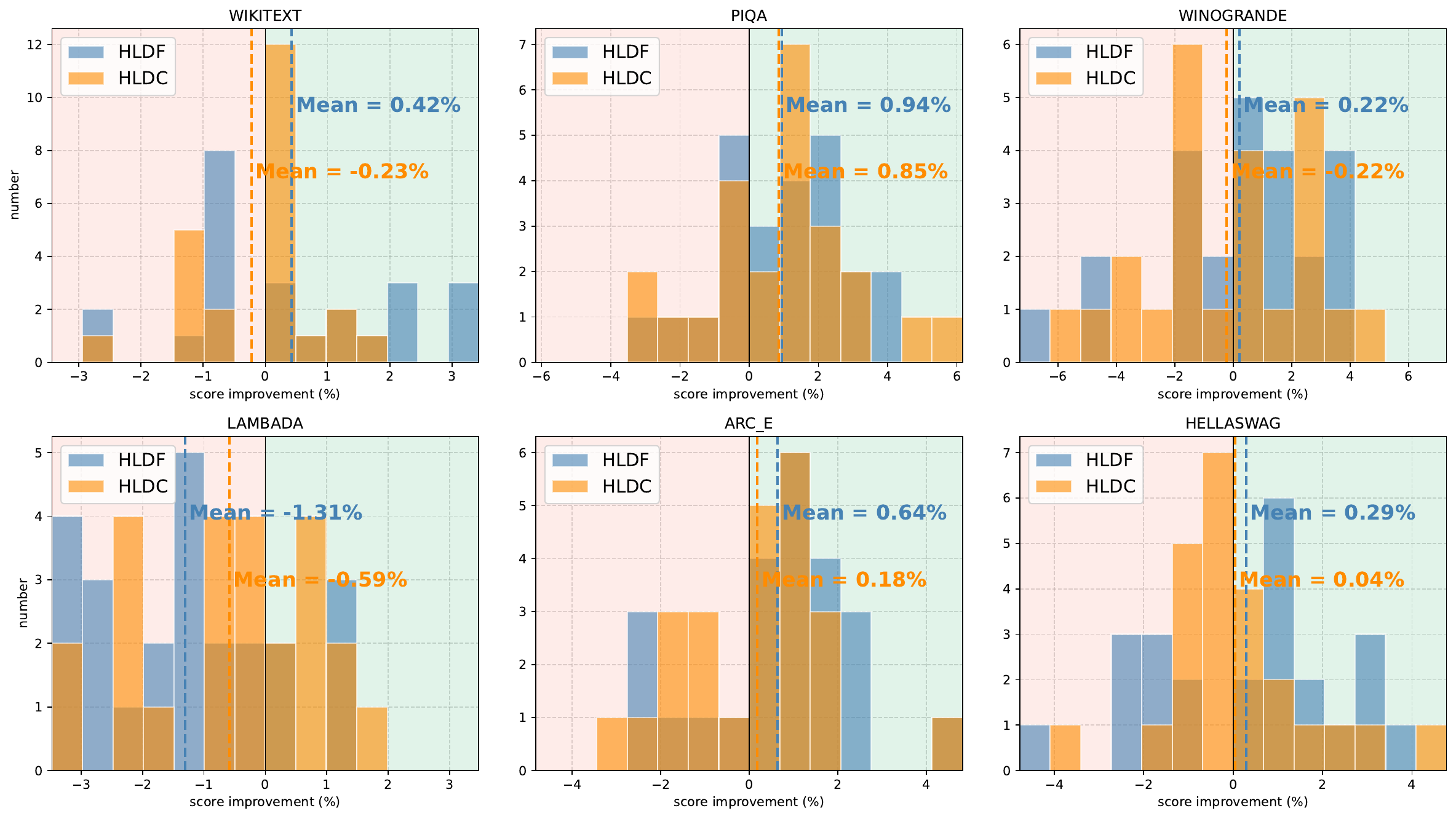}
\caption{
    Shared-HP --- downstream score improvement over KD for the \textbf{735M student}.
    Distribution of pointwise score improvements across all hyperparameter configurations on benchmarks.
}
\label{fig:downstream_735}
\end{figure*}

%% file: plot/downstream_123m.tex
\begin{figure*}[!ht]
    \centering
    \includegraphics[width=\textwidth]{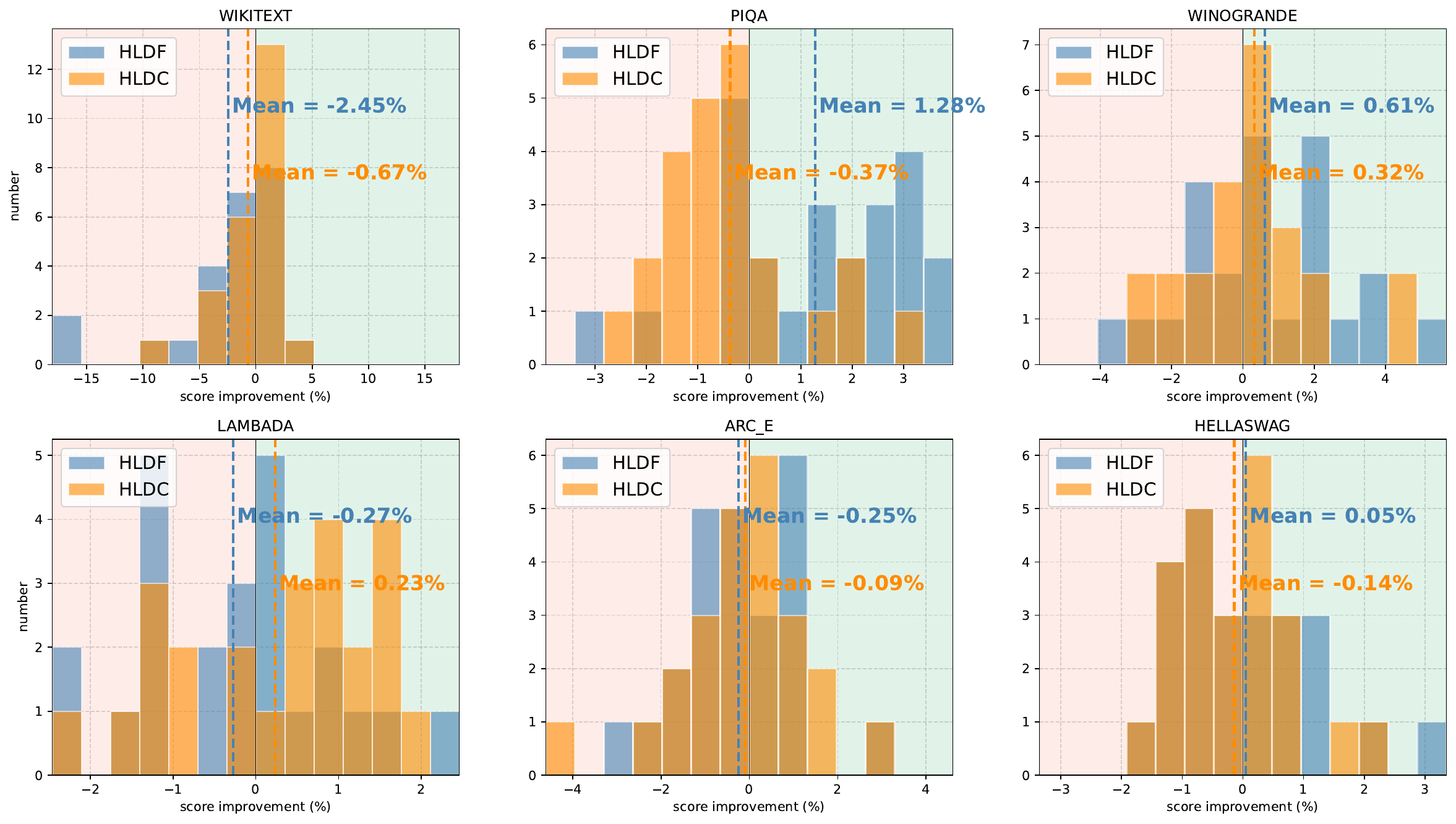}
    \caption{
        Shared-HP --- downstream score improvement over KD for the \textbf{123M student}.
        Distribution of pointwise score improvements across all hyperparameter configurations on benchmarks.
        Green indicates HLD beats KD.
    }
    \label{fig:downstream_123}
\end{figure*}

%% file: appendix.tex
\section{Appendix}\label{sect:appendix}

\subsection{Accounting}
\input{tab/cost_transposed.tex}

\subsection{Scatter plots}
\input{plot/scatter_ppl.tex}
\clearpage
\input{plot/scatter_123m.tex}
\input{plot/scatter_735m.tex}

\clearpage

\subsection{All results}
\input{tab/xps_123.tex}

\clearpage
\input{tab/xps_735.tex}
\clearpage

%% file: tab/cost_transposed.tex
\begin{table*}[h!]
  \centering
  \caption{Compute cost for forward pass, per token.}
      \begin{tabular}{lcccc}
        \toprule
        & Forward pass & Half forward pass & De-embedding & Regressor \\
        \midrule
        Cost & $2N$ & $N$ & $2d_{S}V$ & $2N_{reg}$ \\
        \bottomrule
      \end{tabular}
  \label{tab:cost}
\end{table*}

%% file: plot/scatter_ppl.tex
\begin{figure*}[h!]
    \centering
    \includegraphics[width=\textwidth]{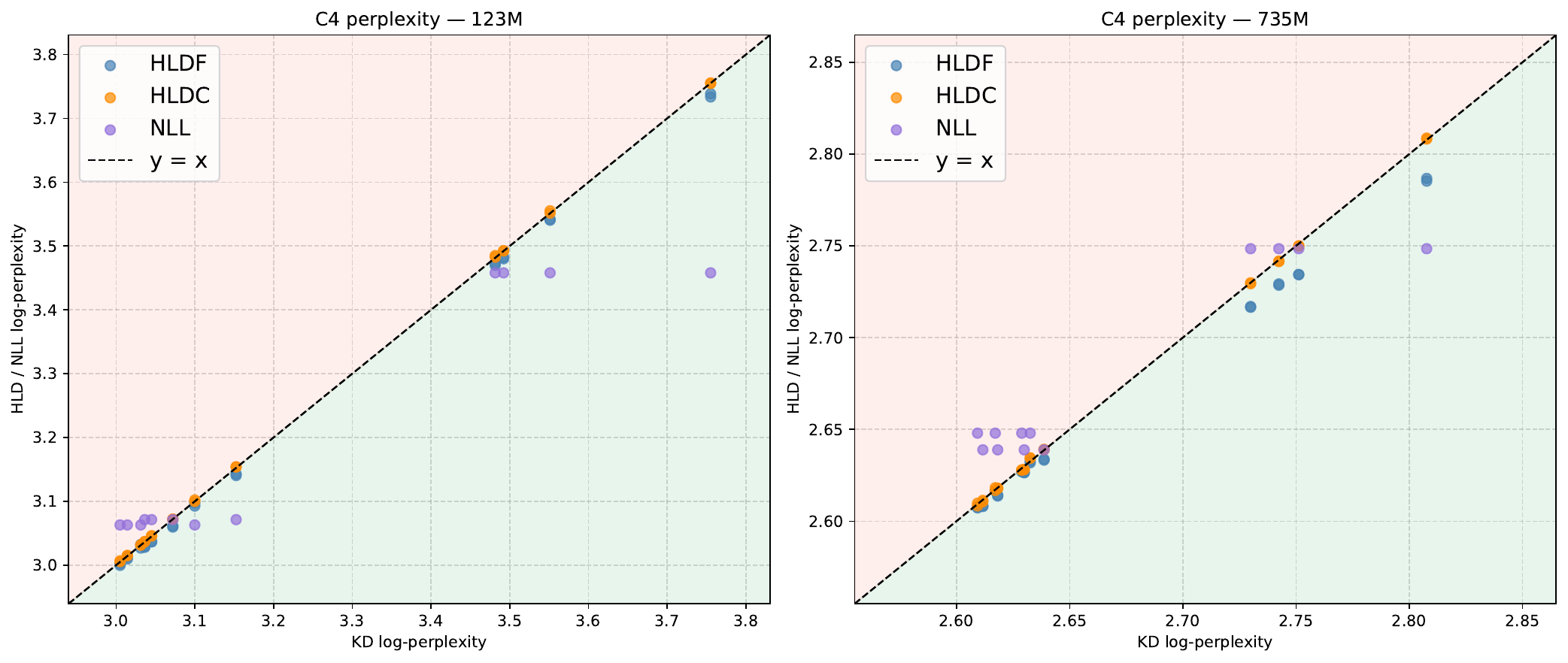}
    \caption{
        C4 perplexity improvement over KD.
        Scatter plot of pointwise score improvements across all hyperparameter configurations on C4 evaluation set. Each point is one hyperparameter set. For each point, the x axis is the KD log-perplexity associated with the hyperparameter set, the y axis is the HLD\&NLL log-perplexity associated with the same hyperparameter set. The lower is always the better so the green zone is where HLD or NLL beats KD.
    }
    \label{fig:scatter_ppl}
\end{figure*}

%% file: plot/scatter_123m.tex
\begin{figure*}[h!]
    \centering
    \includegraphics[width=\textwidth]{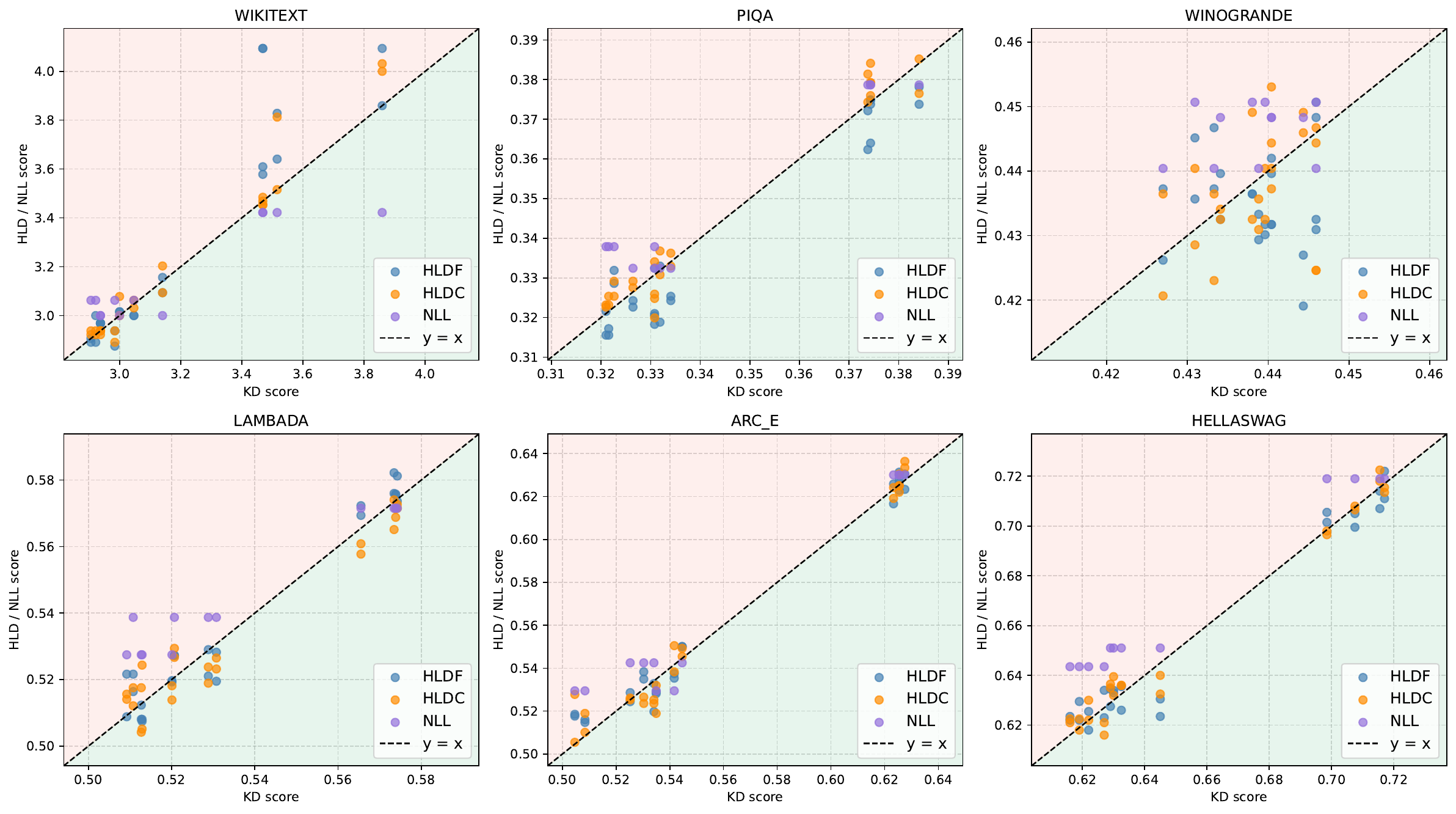}
    \caption{
        Shared-HP --- downstream score improvement over KD for the \textbf{123M student}.
        Scatter plot of pointwise score improvements across all hyperparameter configurations on benchmarks. Each point is one hyperparameter set. For each point, the x axis is the KD score associated with the hyperparameter set, the y axis is the HLD\&NLL score associated with the same hyperparameter set. The lower is always the better so the green zone is where HLD or NLL beats KD.
    }
    \label{fig:scatter_123}
\end{figure*}

%% file: plot/scatter_735m.tex
\begin{figure*}[h!]
    \centering
    \includegraphics[width=\textwidth]{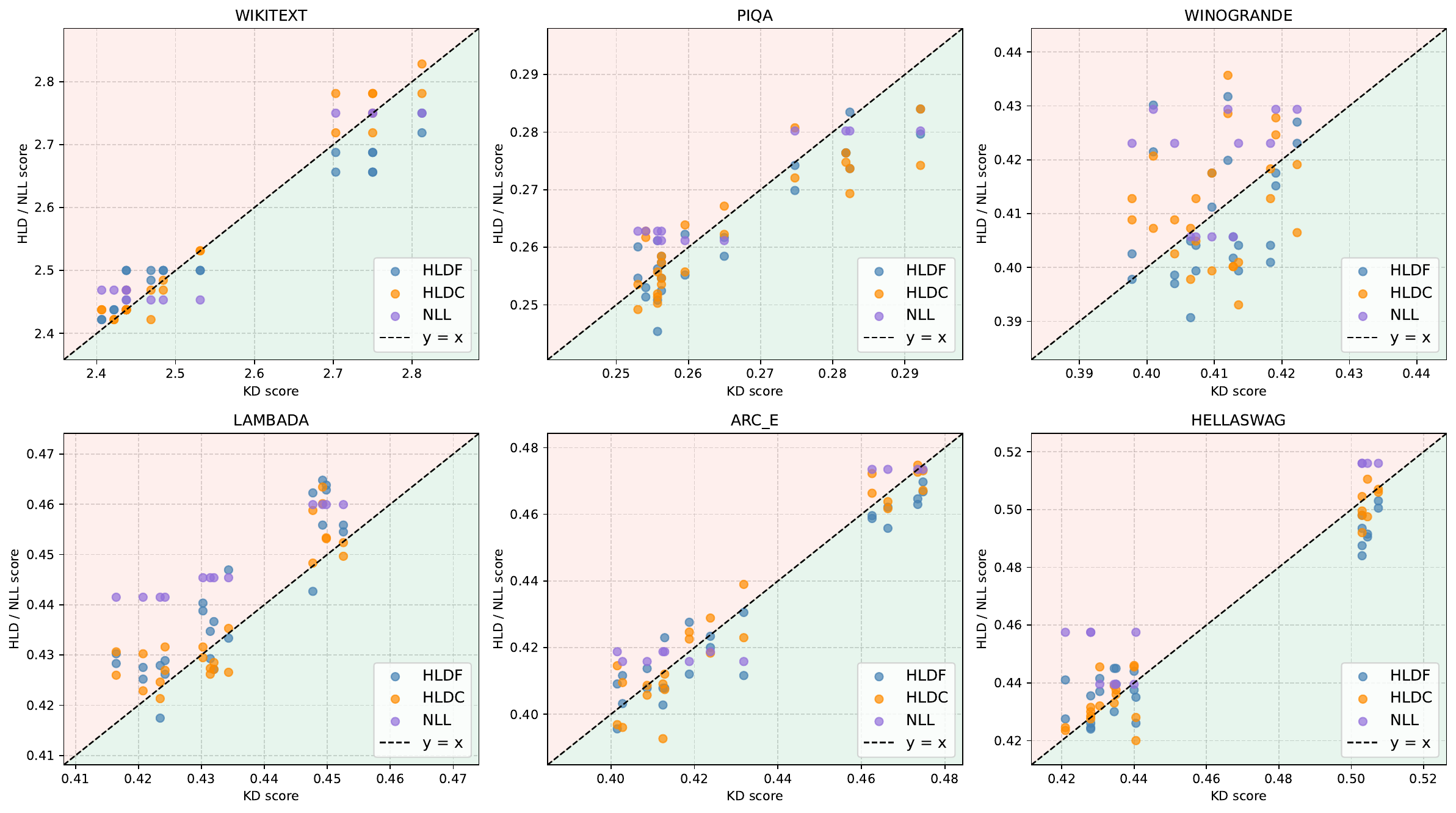}
    \caption{
        Shared-HP --- downstream score improvement over KD for the \textbf{735M student}.
        Scatter plot of pointwise score improvements across all hyperparameter configurations on benchmarks. Each point is one hyperparameter set. For each point, the x axis is the KD score associated with the hyperparameter set, the y axis is the HLD\&NLL score associated with the same hyperparameter set. The lower is always the better so the green zone is where HLD or NLL beats KD.
    }
    \label{fig:scatter_735}
\end{figure*}

%% file: tab/xps_123.tex
\setlength{\LTleft}{-1.5cm}
\setlength{\LTright}{-2cm}
\begin{longtable}{ccccccccccccc}
    \caption{All training runs for the 123M student.}
    \label{tab:xps_123}\\
    \toprule
    \textbf{Method}&$\bm\eta$&$\bm{P_1}$&$\bm\alpha$&$\bm\tau$&$\bm\gamma$&C4&Wikitext&HSwag&Piqa&WinoG&Lambada&Arc-E\\
    \midrule
    \endfirsthead
    \toprule
    \textbf{Method}&$\bm\eta$&$\bm{P_1}$&$\bm\alpha$&$\bm\tau$&$\bm\gamma$&C4&Wikitext&HSwag&Piqa&WinoG&Lambada&Arc-E\\
    \midrule
    \endhead
    NLL&0.0001&0.00&--&--&--&3.458&3.422&0.719&0.379&0.451&0.572&0.630\\
    NLL&0.0004&0.00&--&--&--&3.071&3.000&0.651&0.332&0.448&0.539&0.543\\
    NLL&0.0016&0.00&--&--&--&3.063&3.062&0.643&0.338&0.440&0.527&0.529\\
    KD&0.0001&0.00&0.7&0.5&--&3.551&3.516&0.708&0.374&0.440&0.574&0.625\\
    KD&0.0001&0.00&0.7&1.0&--&3.481&3.469&0.717&0.384&0.438&0.573&0.623\\
    KD&0.0001&0.00&0.9&0.5&--&3.755&3.859&0.699&0.374&0.431&0.565&0.628\\
    KD&0.0001&0.00&0.9&1.0&--&3.492&3.469&0.716&0.374&0.446&0.574&0.625\\
    KD&0.0004&0.00&0.7&0.5&--&3.072&3.000&0.645&0.331&0.444&0.531&0.530\\
    KD&0.0004&0.00&0.7&1.0&--&3.036&2.938&0.630&0.334&0.440&0.521&0.534\\
    KD&0.0004&0.00&0.9&0.5&--&3.152&3.141&0.632&0.332&0.434&0.511&0.545\\
    KD&0.0004&0.00&0.9&1.0&--&3.045&2.938&0.629&0.326&0.440&0.529&0.525\\
    KD&0.0016&0.00&0.7&0.5&--&3.031&2.922&0.627&0.331&0.427&0.520&0.535\\
    KD&0.0016&0.00&0.7&1.0&--&3.005&2.984&0.622&0.321&0.433&0.513&0.505\\
    KD&0.0016&0.00&0.9&0.5&--&3.100&3.047&0.616&0.323&0.446&0.513&0.542\\
    KD&0.0016&0.00&0.9&1.0&--&3.014&2.906&0.619&0.322&0.439&0.509&0.508\\
    HLDC&0.0001&0.00&0.7&0.5&0.05&3.551&3.812&0.708&0.384&0.433&0.571&0.625\\
    HLDC&0.0001&0.00&0.7&0.5&0.10&3.555&3.516&0.707&0.376&0.440&0.569&0.625\\
    HLDC&0.0001&0.00&0.7&1.0&0.05&3.482&3.469&0.716&0.385&0.449&0.574&0.619\\
    HLDC&0.0001&0.00&0.7&1.0&0.10&3.485&3.453&0.714&0.376&0.433&0.565&0.624\\
    HLDC&0.0001&0.00&0.9&0.5&0.05&3.756&4.000&0.697&0.374&0.440&0.561&0.636\\
    HLDC&0.0001&0.00&0.9&0.5&0.10&3.755&4.031&0.698&0.381&0.429&0.558&0.633\\
    HLDC&0.0001&0.00&0.9&1.0&0.05&3.493&3.484&0.723&0.379&0.447&0.573&0.630\\
    HLDC&0.0001&0.00&0.9&1.0&0.10&3.492&3.453&0.718&0.379&0.444&0.572&0.622\\
    HLDC&0.0004&0.00&0.7&0.5&0.05&3.072&3.078&0.640&0.326&0.446&0.523&0.524\\
    HLDC&0.0004&0.00&0.7&0.5&0.10&3.072&3.000&0.632&0.334&0.449&0.526&0.527\\
    HLDC&0.0004&0.00&0.7&1.0&0.05&3.037&2.922&0.639&0.333&0.440&0.529&0.524\\
    HLDC&0.0004&0.00&0.7&1.0&0.10&3.037&2.938&0.632&0.336&0.437&0.527&0.525\\
    HLDC&0.0004&0.00&0.9&0.5&0.05&3.154&3.094&0.636&0.331&0.433&0.512&0.549\\
    HLDC&0.0004&0.00&0.9&0.5&0.10&3.154&3.203&0.636&0.337&0.434&0.518&0.545\\
    HLDC&0.0004&0.00&0.9&1.0&0.05&3.046&2.938&0.636&0.329&0.453&0.524&0.525\\
    HLDC&0.0004&0.00&0.9&1.0&0.10&3.045&2.938&0.635&0.328&0.444&0.519&0.526\\
    HLDC&0.0016&0.00&0.7&0.5&0.05&3.032&2.938&0.616&0.320&0.436&0.514&0.519\\
    HLDC&0.0016&0.00&0.7&0.5&0.10&3.031&2.922&0.621&0.325&0.421&0.518&0.532\\
    HLDC&0.0016&0.00&0.7&1.0&0.05&3.005&2.938&0.630&0.323&0.423&0.505&0.505\\
    HLDC&0.0016&0.00&0.7&1.0&0.10&3.007&2.891&0.622&0.323&0.436&0.524&0.528\\
    HLDC&0.0016&0.00&0.9&0.5&0.05&3.100&3.031&0.623&0.325&0.425&0.504&0.538\\
    HLDC&0.0016&0.00&0.9&0.5&0.10&3.102&3.062&0.621&0.329&0.425&0.518&0.551\\
    HLDC&0.0016&0.00&0.9&1.0&0.05&3.015&2.922&0.618&0.323&0.431&0.516&0.519\\
    HLDC&0.0016&0.00&0.9&1.0&0.10&3.015&2.938&0.623&0.325&0.436&0.514&0.510\\
    HLDF&0.0001&0.01&0.7&0.5&--&3.540&3.828&0.705&0.375&0.430&0.576&0.628\\
    HLDF&0.0001&0.05&0.7&0.5&--&3.542&3.641&0.700&0.364&0.432&0.576&0.623\\
    HLDF&0.0001&0.01&0.7&1.0&--&3.473&4.094&0.722&0.378&0.436&0.576&0.617\\
    HLDF&0.0001&0.05&0.7&1.0&--&3.470&3.578&0.711&0.374&0.436&0.582&0.626\\
    HLDF&0.0001&0.01&0.9&0.5&--&3.739&4.094&0.706&0.372&0.436&0.569&0.623\\
    HLDF&0.0001&0.05&0.9&0.5&--&3.733&3.859&0.702&0.362&0.445&0.572&0.630\\
    HLDF&0.0001&0.01&0.9&1.0&--&3.482&4.094&0.707&0.375&0.448&0.581&0.631\\
    HLDF&0.0001&0.05&0.9&1.0&--&3.480&3.609&0.714&0.374&0.451&0.574&0.626\\
    HLDF&0.0004&0.01&0.7&0.5&--&3.061&3.016&0.630&0.320&0.427&0.528&0.535\\
    HLDF&0.0004&0.05&0.7&0.5&--&3.060&3.016&0.624&0.332&0.419&0.520&0.538\\
    HLDF&0.0004&0.01&0.7&1.0&--&3.029&2.969&0.633&0.324&0.442&0.527&0.520\\
    HLDF&0.0004&0.05&0.7&1.0&--&3.028&2.969&0.634&0.325&0.432&0.527&0.533\\
    HLDF&0.0004&0.01&0.9&0.5&--&3.143&3.156&0.635&0.333&0.440&0.522&0.550\\
    HLDF&0.0004&0.05&0.9&0.5&--&3.140&3.094&0.626&0.319&0.433&0.516&0.550\\
    HLDF&0.0004&0.01&0.9&1.0&--&3.037&2.969&0.634&0.323&0.440&0.529&0.524\\
    HLDF&0.0004&0.05&0.9&1.0&--&3.036&2.969&0.627&0.324&0.432&0.521&0.529\\
    HLDF&0.0016&0.01&0.7&0.5&--&3.032&3.000&0.623&0.321&0.426&0.519&0.528\\
    HLDF&0.0016&0.05&0.7&0.5&--&3.027&2.891&0.634&0.318&0.437&0.520&0.529\\
    HLDF&0.0016&0.01&0.7&1.0&--&3.002&2.938&0.618&0.316&0.447&0.508&0.518\\
    HLDF&0.0016&0.05&0.7&1.0&--&3.000&2.875&0.625&0.322&0.437&0.507&0.519\\
    HLDF&0.0016&0.01&0.9&0.5&--&3.097&3.000&0.624&0.332&0.431&0.512&0.535\\
    HLDF&0.0016&0.05&0.9&0.5&--&3.093&3.000&0.622&0.329&0.433&0.508&0.537\\
    HLDF&0.0016&0.01&0.9&1.0&--&3.012&2.906&0.622&0.317&0.433&0.522&0.516\\
    HLDF&0.0016&0.05&0.9&1.0&--&3.010&2.891&0.629&0.316&0.429&0.509&0.515\\
    \bottomrule
\end{longtable}

%% file: tab/xps_735.tex
\begin{longtable}{ccccccccccccc}
    \caption{All training runs for the 735M student.}\label{tab:xps_735}\\
    \toprule
    \textbf{Method}&$\bm\eta$&$\bm{P_1}$&$\bm\alpha$&$\bm\tau$&$\bm\gamma$&C4&Wikitext&HSwag&Piqa&WinoG&Lambada&Arc-E\\
    \midrule
    \endfirsthead
    \toprule
    \textbf{Method}&$\bm\eta$&$\bm{P_1}$&$\bm\alpha$&$\bm\tau$&$\bm\gamma$&C4&Wikitext&HellaSwag&Piqa&WinoGrande&Lambada&Arc-E\\
    \midrule
    \endhead
    NLL&0.0001&0.00&--&--&--&2.748&2.750&0.516&0.280&0.429&0.460&0.473\\
    NLL&0.0004&0.00&--&--&--&2.639&2.453&0.440&0.261&0.423&0.445&0.416\\
    NLL&0.0016&0.00&--&--&--&2.648&2.469&0.458&0.263&0.406&0.441&0.419\\
    KD&0.0001&0.00&0.7&0.5&--&2.751&2.750&0.503&0.282&0.419&0.448&0.475\\
    KD&0.0001&0.00&0.7&1.0&--&2.730&2.750&0.504&0.282&0.412&0.450&0.466\\
    KD&0.0001&0.00&0.9&0.5&--&2.808&2.812&0.503&0.292&0.401&0.453&0.473\\
    KD&0.0001&0.00&0.9&1.0&--&2.742&2.703&0.507&0.275&0.422&0.449&0.463\\
    KD&0.0004&0.00&0.7&0.5&--&2.618&2.484&0.430&0.260&0.398&0.434&0.419\\
    KD&0.0004&0.00&0.7&1.0&--&2.612&2.469&0.434&0.256&0.404&0.430&0.409\\
    KD&0.0004&0.00&0.9&0.5&--&2.639&2.531&0.435&0.265&0.414&0.431&0.432\\
    KD&0.0004&0.00&0.9&1.0&--&2.630&2.438&0.440&0.256&0.418&0.432&0.403\\
    KD&0.0016&0.00&0.7&0.5&--&2.617&2.422&0.441&0.254&0.413&0.424&0.413\\
    KD&0.0016&0.00&0.7&1.0&--&2.609&2.406&0.428&0.256&0.407&0.423&0.402\\
    KD&0.0016&0.00&0.9&0.5&--&2.633&2.438&0.428&0.256&0.406&0.421&0.424\\
    KD&0.0016&0.00&0.9&1.0&--&2.629&2.438&0.421&0.253&0.410&0.416&0.412\\
    HLDC&0.0001&0.00&0.7&0.5&0.05&2.750&2.781&0.498&0.276&0.428&0.459&0.467\\
    HLDC&0.0001&0.00&0.7&0.5&0.10&2.750&2.781&0.504&0.275&0.425&0.448&0.473\\
    HLDC&0.0001&0.00&0.7&1.0&0.05&2.729&2.719&0.497&0.274&0.429&0.453&0.464\\
    HLDC&0.0001&0.00&0.7&1.0&0.10&2.730&2.750&0.510&0.269&0.436&0.453&0.462\\
    HLDC&0.0001&0.00&0.9&0.5&0.05&2.809&2.781&0.499&0.274&0.407&0.450&0.473\\
    HLDC&0.0001&0.00&0.9&0.5&0.10&2.808&2.828&0.492&0.284&0.421&0.452&0.475\\
    HLDC&0.0001&0.00&0.9&1.0&0.05&2.742&2.719&0.507&0.272&0.406&0.463&0.472\\
    HLDC&0.0001&0.00&0.9&1.0&0.10&2.741&2.781&0.506&0.281&0.419&0.460&0.466\\
    HLDC&0.0004&0.00&0.7&0.5&0.05&2.618&2.484&0.446&0.264&0.413&0.427&0.423\\
    HLDC&0.0004&0.00&0.7&0.5&0.10&2.618&2.469&0.432&0.256&0.409&0.435&0.425\\
    HLDC&0.0004&0.00&0.7&1.0&0.05&2.611&2.422&0.439&0.252&0.403&0.432&0.406\\
    HLDC&0.0004&0.00&0.7&1.0&0.10&2.611&2.469&0.433&0.256&0.409&0.429&0.409\\
    HLDC&0.0004&0.00&0.9&0.5&0.05&2.639&2.531&0.438&0.267&0.401&0.427&0.423\\
    HLDC&0.0004&0.00&0.9&0.5&0.10&2.639&2.531&0.436&0.262&0.393&0.426&0.439\\
    HLDC&0.0004&0.00&0.9&1.0&0.05&2.628&2.438&0.446&0.255&0.413&0.427&0.410\\
    HLDC&0.0004&0.00&0.9&1.0&0.10&2.628&2.469&0.446&0.254&0.418&0.428&0.396\\
    HLDC&0.0016&0.00&0.7&0.5&0.05&2.617&2.422&0.420&0.263&0.400&0.432&0.407\\
    HLDC&0.0016&0.00&0.7&0.5&0.10&2.618&2.422&0.428&0.262&0.400&0.427&0.412\\
    HLDC&0.0016&0.00&0.7&1.0&0.05&2.608&2.438&0.430&0.251&0.405&0.425&0.415\\
    HLDC&0.0016&0.00&0.7&1.0&0.10&2.610&2.438&0.431&0.250&0.413&0.421&0.397\\
    HLDC&0.0016&0.00&0.9&0.5&0.05&2.634&2.438&0.428&0.257&0.407&0.423&0.429\\
    HLDC&0.0016&0.00&0.9&0.5&0.10&2.635&2.438&0.427&0.258&0.398&0.430&0.418\\
    HLDC&0.0016&0.00&0.9&1.0&0.05&2.628&2.438&0.423&0.249&0.418&0.426&0.393\\
    HLDC&0.0016&0.00&0.9&1.0&0.10&2.628&2.438&0.424&0.254&0.399&0.431&0.409\\
    HLDF&0.0001&0.01&0.7&0.5&--&2.734&2.656&0.498&0.276&0.415&0.443&0.467\\
    HLDF&0.0001&0.04&0.7&0.5&--&2.734&2.688&0.493&0.276&0.418&0.462&0.470\\
    HLDF&0.0001&0.01&0.7&1.0&--&2.716&2.656&0.490&0.283&0.420&0.463&0.462\\
    HLDF&0.0001&0.04&0.7&1.0&--&2.717&2.688&0.491&0.274&0.432&0.464&0.456\\
    HLDF&0.0001&0.01&0.9&0.5&--&2.787&2.750&0.487&0.284&0.430&0.454&0.463\\
    HLDF&0.0001&0.04&0.9&0.5&--&2.785&2.719&0.484&0.280&0.421&0.456&0.465\\
    HLDF&0.0001&0.01&0.9&1.0&--&2.729&2.688&0.500&0.270&0.427&0.456&0.460\\
    HLDF&0.0001&0.04&0.9&1.0&--&2.729&2.656&0.503&0.274&0.423&0.465&0.459\\
    HLDF&0.0004&0.01&0.7&0.5&--&2.614&2.500&0.442&0.262&0.398&0.433&0.428\\
    HLDF&0.0004&0.04&0.7&0.5&--&2.614&2.500&0.437&0.255&0.403&0.447&0.412\\
    HLDF&0.0004&0.01&0.7&1.0&--&2.608&2.500&0.430&0.256&0.399&0.439&0.414\\
    HLDF&0.0004&0.04&0.7&1.0&--&2.608&2.484&0.445&0.251&0.397&0.440&0.408\\
    HLDF&0.0004&0.01&0.9&0.5&--&2.634&2.500&0.440&0.258&0.399&0.429&0.431\\
    HLDF&0.0004&0.04&0.9&0.5&--&2.633&2.500&0.445&0.262&0.404&0.435&0.412\\
    HLDF&0.0004&0.01&0.9&1.0&--&2.626&2.500&0.438&0.255&0.401&0.427&0.412\\
    HLDF&0.0004&0.04&0.9&1.0&--&2.626&2.500&0.444&0.252&0.404&0.437&0.403\\
    HLDF&0.0016&0.01&0.7&0.5&--&2.617&2.438&0.426&0.251&0.402&0.429&0.423\\
    HLDF&0.0016&0.04&0.7&0.5&--&2.617&2.438&0.435&0.253&0.406&0.426&0.408\\
    HLDF&0.0016&0.01&0.7&1.0&--&2.607&2.422&0.426&0.261&0.404&0.428&0.409\\
    HLDF&0.0016&0.04&0.7&1.0&--&2.608&2.422&0.424&0.245&0.399&0.417&0.396\\
    HLDF&0.0016&0.01&0.9&0.5&--&2.633&2.453&0.435&0.257&0.391&0.428&0.420\\
    HLDF&0.0016&0.04&0.9&0.5&--&2.632&2.438&0.424&0.258&0.405&0.425&0.423\\
    HLDF&0.0016&0.01&0.9&1.0&--&2.627&2.438&0.427&0.255&0.418&0.430&0.403\\
    HLDF&0.0016&0.04&0.9&1.0&--&2.627&2.438&0.441&0.260&0.411&0.428&0.408\\
    \bottomrule
\end{longtable}